\documentclass[conference]{IEEEtran}
\IEEEoverridecommandlockouts

\newif\ifreview

\usepackage{cite}
\usepackage{amsmath,amssymb,amsfonts}
\usepackage{algorithmic}
\usepackage{graphicx}
\usepackage{textcomp}
\usepackage{xcolor}
\def\BibTeX{{\rm B\kern-.05em{\sc i\kern-.025em b}\kern-.08em
    T\kern-.1667em\lower.7ex\hbox{E}\kern-.125emX}}

\usepackage{diagbox}
\usepackage{pifont}
\usepackage{makecell}
\usepackage{array}   
\usepackage{multirow}
\usepackage{booktabs}
\usepackage{hyperref}
    
\begin{document}

\title{Adapting Foundation ASR Models to Dysarthric Speech: A Case Study
}

\ifreview
\author{
  \IEEEauthorblockN{Anonymous Authors}
  \IEEEauthorblockA{
    \textit{Anonymous Institution} \\
    City, Country \\
    anonymous@institution.example
  }
}
\else
\author{\IEEEauthorblockN{1\textsuperscript{st} Christian Huber}
\IEEEauthorblockA{\textit{Interactive Systems Lab} \\
\textit{Karlsruhe Institute of Technology}\\
Karlsruhe, Germany\\
christian.huber@kit.edu}
\and
\IEEEauthorblockN{2\textsuperscript{nd} Laura Kernahan}
\IEEEauthorblockA{\textit{Interactive Systems Lab} \\
\textit{Karlsruhe Institute of Technology}\\
Karlsruhe, Germany\\
laura.kernahan@student.kit.edu}
\and
\IEEEauthorblockN{3\textsuperscript{rd} Alexander Waibel}
\IEEEauthorblockA{\textit{Interactive Systems Lab} \\
\textit{Carnegie Mellon University}\\
Pittsburgh PA, USA\\
alexander.waibel@cmu.edu}
}
\fi

\maketitle

\begin{abstract}
Automatic speech recognition (ASR) systems often perform poorly in dysarthric speech, limiting their usefulness to affected speakers in everyday communication. This paper presents a personalized ASR system for a dysarthric speaker, built by adapting a foundation ASR model to speaker-specific data. Using the TEQST tool, we collected 92 hours of read speech and later added 8.8 hours of user corrections gathered through a deployed mobile application. Starting from Whisper, fine-tuning reduced word error rate to 15.8\% with only 1.4 hours of adaptation data on the read test set, reached 10.7\% / 16.1\% with 22.5 hours on the read and corrections test sets, respectively, and achieved the best result of 9.7\% when using all available data including the corrections on the read test set and 7.8\% on the corrections test set. Using LoRA adaptation and/or Qwen3-ASR as foundation model performed worse in this setting. The results show that personalized fine-tuning can make foundation ASR models substantially more effective for dysarthric speech and suitable for practical deployment.
\end{abstract}

\begin{IEEEkeywords}
automatic speech recognition, dysarthric speech
\end{IEEEkeywords}

\section{Introduction}

ASR has advanced rapidly\cite{vaswani2017attention,pham2019very},
especially with the development of large foundation models\cite{radford2023robust,Qwen3-ASR}, which can achieve near-human performance on standard speech benchmarks.
However, these models are mostly trained on standard speech and perform poorly, when confronted with dysarthric speech. This excludes users who could benefit immensely from this technology. 
The German speaker in this study is affected by the John Cunningham virus (JC-virus),
which can cause Progressive Multifocal Leukoencephalopathy (PML), 
a serious brain disease caused by demyelination of white matter in the central nervous system. This neurological damage affects motor control including speech, resulting in dysarthria. A sample video of the dysarthric speech which shows the severity of the disease can be seen
\ifreview
\href{BLIND}{here (BLIND)}.
\else
\href{https://lecture-translator.kit.edu/archivesession/Ly9PdGhlci9Eb2VocmluZy9Eb2VocmluZ1NwcmVjaHByb2JlMlZpZGVv}{here}.
\fi
The tester's speech presents significant challenges for a standard ASR system, as evidenced by the baseline performance (see Section \ref{sec:exp}). 
The same is true for humans and it requires a lot of concentration when trying to understand the speaker.
Despite the named challenges, speech remains the preferred communication modality of our test user, as it enables natural face-to-face interaction.

The main contributions of this paper are:
1) a personalised ASR pipeline for a dysarthric speaker achieving 9.7\% / 7.8\% WER using a fine-tuned Whisper model,
2) a comparative evaluation of how performance depends on the amount of adaptation data, full fine-tuning versus LoRA adaptation, and Whisper versus Qwen3-ASR, and
3) a deployed mobile application enabling continuous data collection through real-world usage. 

\section{Related Work}
\subsection{Foundation ASR Models}
Recent advances in self-supervised and weakly-supervised pretraining have produced foundation ASR models such as Whisper \cite{radford2023robust}, trained on large-scale weakly labeled data, and more recently Qwen3-ASR \cite{Qwen3-ASR}, which leverages a large language model backbone for speech understanding.

\subsection{Dysarthric Speech Recognition}
Recognizing dysarthric speech remains a long-standing challenge due to its acoustic variability and the scarcity of available training data \cite{strik2025speech}. Early approaches relied on speaker-independent acoustic models combined with data augmentation strategies, including speed perturbation and GAN-based adversarial augmentation \cite{wang2024enhancing}, voice conversion \cite{towards_inclusive_2025}, and synthetic data generation via personalized text-to-speech \cite{improved_dysarthric_tts2025}.

\begin{figure*}[t]
  \centering
  \includegraphics[trim=0.2cm 0.2cm 0.2cm 0.2cm,clip,width=1\textwidth]{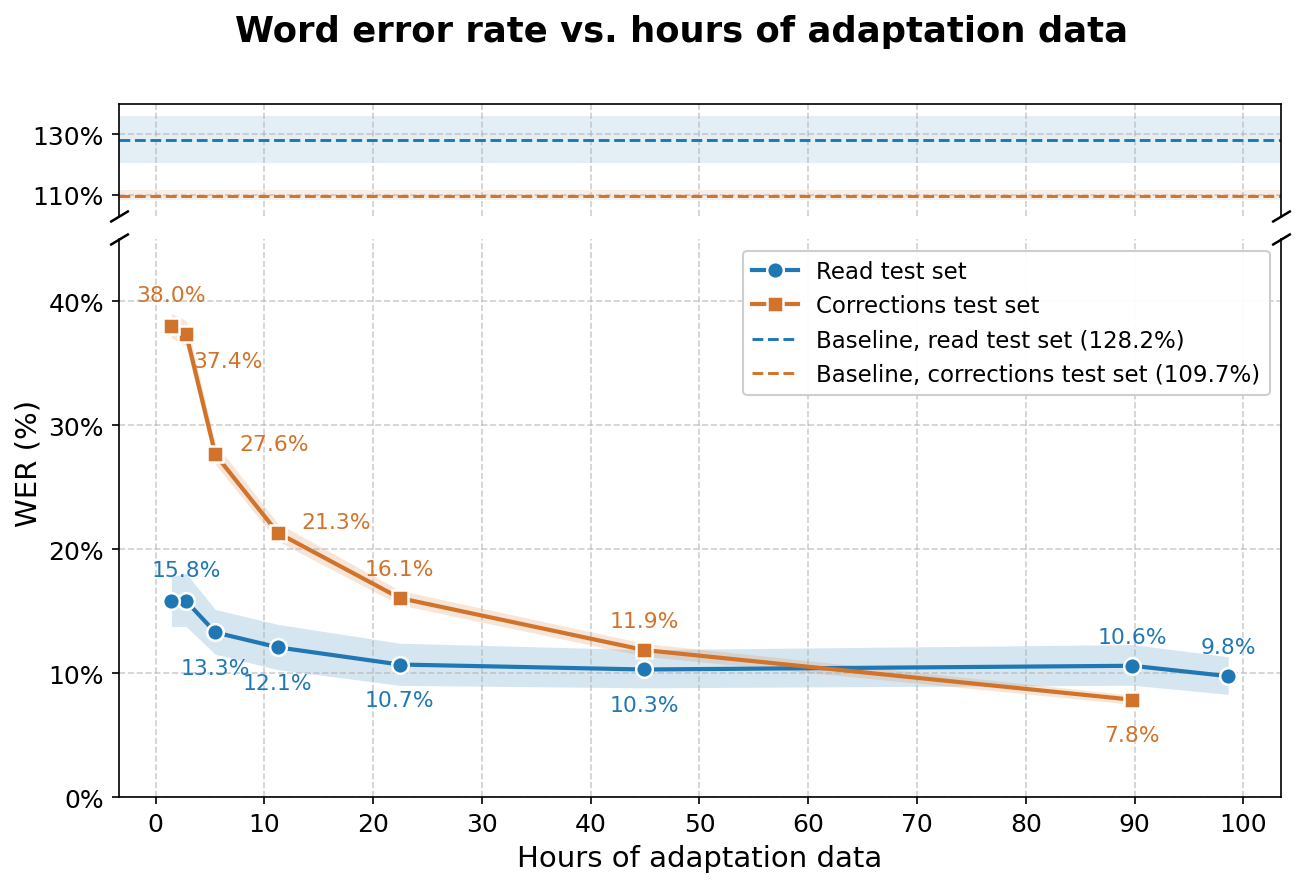}
  \caption{WER on the test sets vs. hours of dysarthric adaptation data. Shaded bands: 95\% bootstrap confidence intervals (1000 resamples). The 98.6 h model is omitted from the corrections curve — the corrections data was seen during its training.}
  \label{fig:wer}
\end{figure*}

More closely related to our work, several studies have demonstrated that personalized fine-tuning on speaker-specific dysarthric speech substantially outperforms speaker-independent models \cite{green2019personalizing}. For instance, personalized fine-tuning has achieved WER reductions from 62\% to 35\% relative improvement for speakers with amyotrophic lateral sclerosis (ALS), reaching 10\% WER for mild and 20\% WER for more severe dysarthria \cite{green2019personalizing}. Other work has explored on-the-fly personalization via in-context learning to avoid the cost of training individual models per speaker \cite{metaicl2025}.

Our work differs from prior studies in three ways: we evaluate both full fine-tuning and parameter-efficient LoRA adaptation \cite{hu2022lora} on a single speaker with a substantially larger adaptation dataset (up to 100 hours); we compare two different foundation model families (Whisper and Qwen3-ASR); and we deploy the resulting system in a real-world mobile application that enables continuous data collection through user corrections, going beyond a purely offline evaluation.

\begin{table}[t]
\centering
\caption{Dataset statistics.}
\label{tab:data}
\begin{tabular}{lrr}
\toprule
\textbf{Split} & \textbf{Utterances} & \textbf{Duration (h)} \\
\midrule
Train            & 22{,}572 &  89.8 \\
Corrections      & 4{,}446  &   8.8 \\
Dev              &    209   &   1.1 \\
Test             &    217   &   1.1 \\
\midrule
\textbf{Total}   & \textbf{27{,}444} & \textbf{100.8} \\
\bottomrule
\end{tabular}
\end{table}

\section{Data}

To collect the dysarthric speech data, we used the TEQST tool \cite{teqst}. The data set can be found 
\ifreview
\href{BLIND}{here (BLIND)}.
\else
\href{https://huggingface.co/datasets/chuber/dysarthric-speech}{here}.
\fi
The tool enabled the user to record the data from home without supervision. The user explained that collecting the data otherwise would have been impossible, because recording took a lot of effort with considerable breaks.
One speaker read 92 hours, which we divided into training, development, and test sets (see Table \ref{tab:data}). The first 1.1h the speaker read were used for the test set and the second 1.1h were used for the development set. The text data provided was from the fairy tale domain.
We adapted a model on this data (see Section \ref{sec:exp}) and deployed it in a mobile app (see Section \ref{sec:deployment}). The speaker was using the model in everyday life and could correct errors through the app. With this, we collected another 8.8 hours of data.

\section{Experiments and Results}
\label{sec:exp}

We started with Whisper \cite{radford2023robust} (whisper-large-v3) as a baseline model and compared fine-tuning the full model and adapted with added LoRA \cite{hu2022lora} weights only.

The fine-tuning results can be seen in Figure \ref{fig:wer}. Using only 1.4h of adaptation data, we observe on the read test set a WER of 15.8\%, which is already a huge improvement over the 128.4\% WER of the baseline model. The same holds for the corrections test set where the WER is 38.0\% compared to 109.7\% of the baseline model. With 22.5h of adaptation data we get a 10.7\% and 16.1\% WER, respectively, which is a good trade off between data amount and performance. Adding all the data including the corrections performed the best with 9.7\% WER. Using all read data scored 7.8\% WER on the corrections test set.

Adapting LoRA weights only (with rank 8) performed substantially worse (between 15\% and 39\% relative) for the same amount of adaptation data and falls outside the specified confidence intervals for the fully fine-tuned model. The reason for that may be that dysarthric speech is very different from speech the baseline model was trained on and the added LoRA weights did not have the ability to learn this difference.

Furthermore, we performed the same experiments with Qwen3-ASR-1.7B \cite{Qwen3-ASR} as a baseline model. However, the performance was worse than that of Whisper. Using all the adaptation data, we obtained a WER between 14\% and 16\% on the read test set for fine-tuning and LoRA, respectively. This also falls outside the confidence intervals of the models based on Whisper.

\section{Deployment}
\label{sec:deployment}

\begin{figure}[t]
  \centering
  \boxed{\includegraphics[width=0.29\columnwidth]{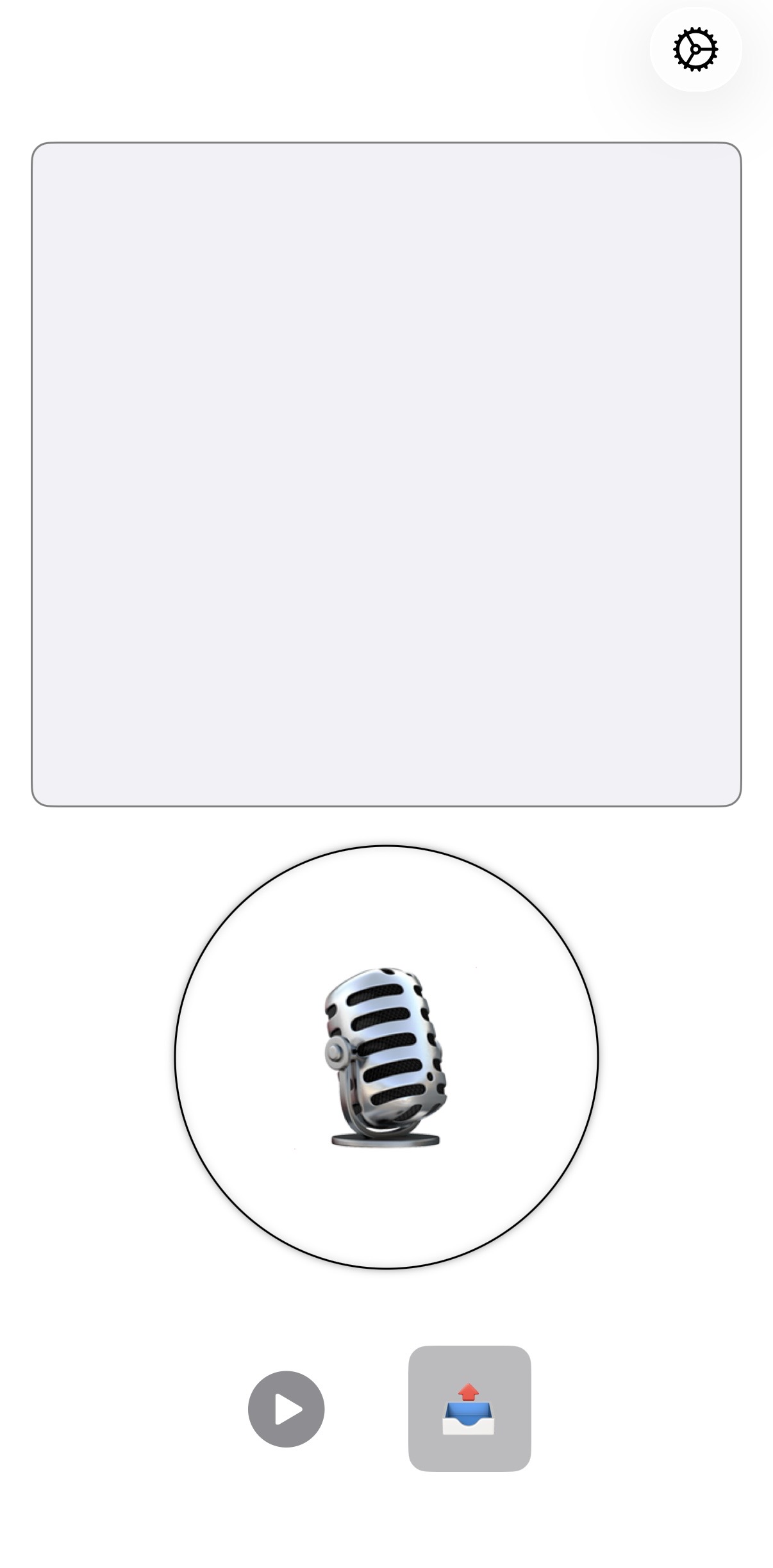}}
  \boxed{\includegraphics[width=0.29\columnwidth,trim=0cm 0cm 0cm 0.5cm,clip]{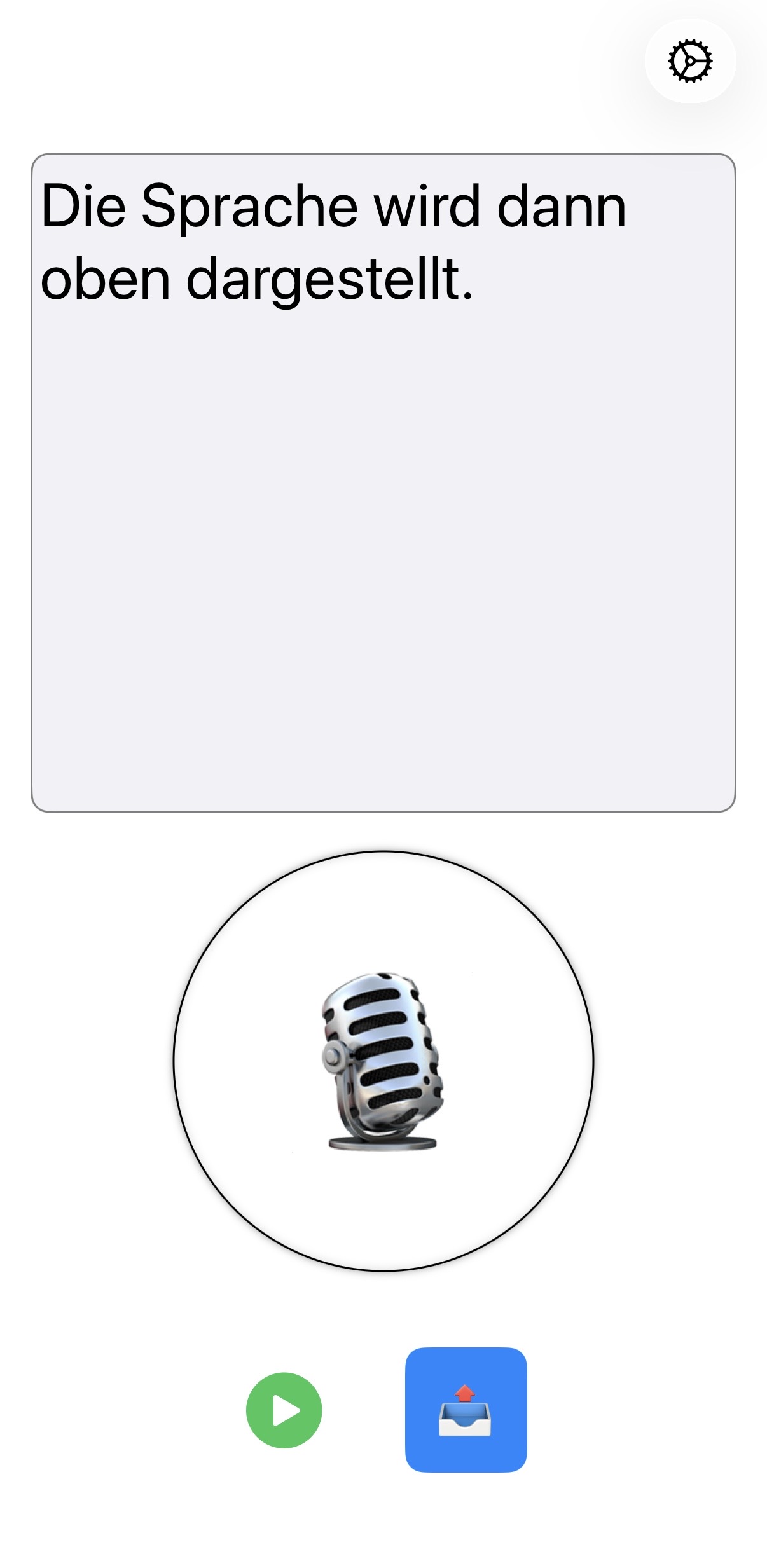}}
  \boxed{\includegraphics[width=0.29\columnwidth]{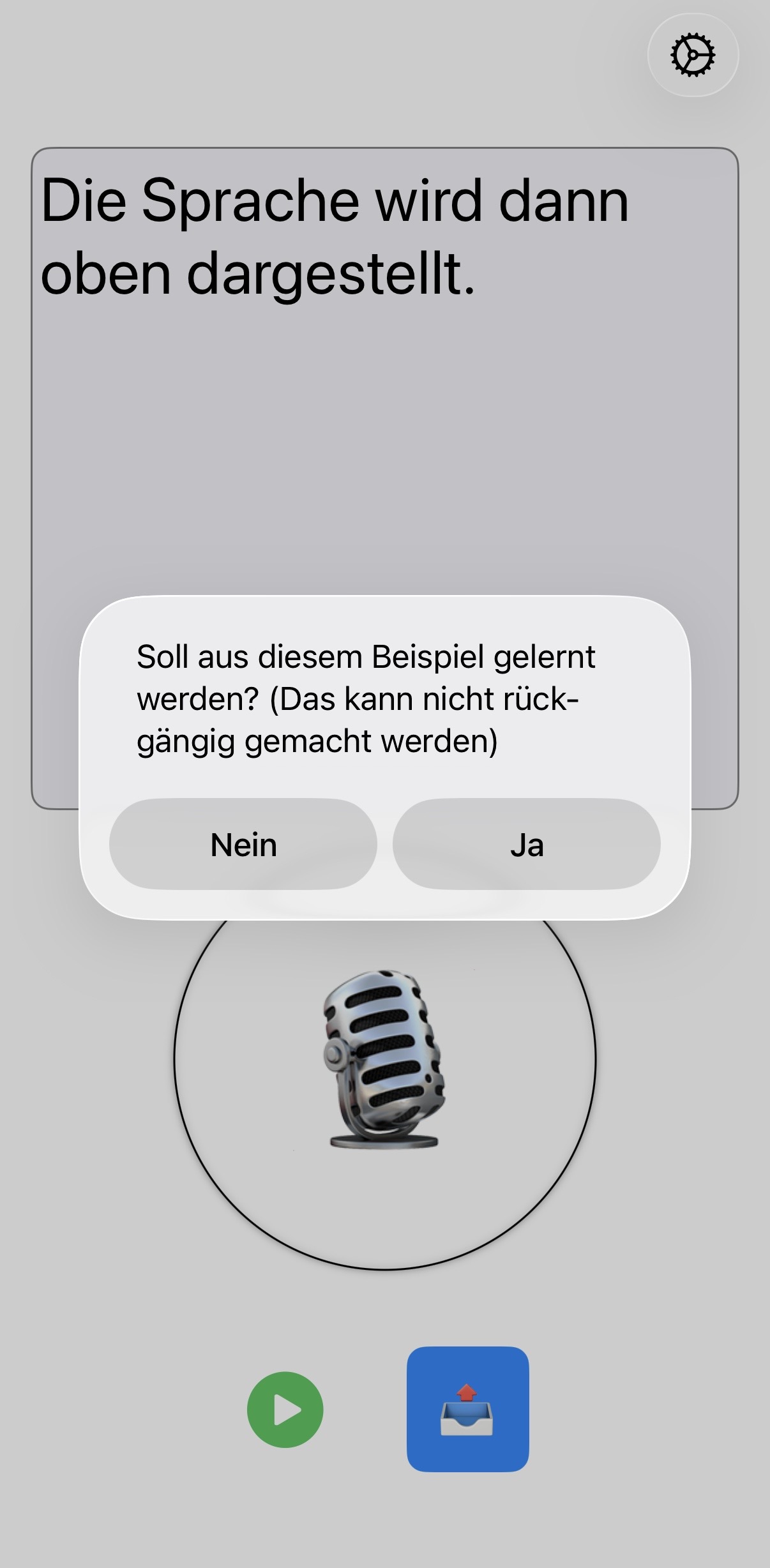}}
  \caption{The iOS app: (left) main recording interface, 
  (center) transcription displayed after recording, 
  (right) correction confirmation dialog for continuous model improvement.}
  \label{fig:app}
\end{figure}

The fine-tuned model was deployed, first as an android application and later as an iOS app. The iOS deployment functioned as a personalized speech communicator on the test user phone (see Figure \ref{fig:app}). It is designed around the specific needs of the dysarthric speaker, prioritizing accessibility, simplicity and a natural communication experience.

The model is hosted in the cloud using the faster whisper\cite{faster_whisper} implementation which is based on CTranslate2\cite{klein2020opennmt,ctranslate2}.
Audio recorded on the device is transmitted to the server and the resulting transcription is returned to the app in real time. The server-side approach was used for fast inference without placing computational demands on the phone, resulting in much faster transcription. 

The app features a minimal, high-contrast interface cantered around a single large microphone button, making it easy to operate with limited fine motor control (see Figure \ref{fig:app}). The transcribed text is displayed in a large text box above the microphone button (see Figure \ref{fig:app}). 

The speaker has been included in the development and has contributed significantly to the app setup, making the app ideal for motor- and speech-impaired users.

We have provided different interaction options to accommodate different speaker preferences. Three recording modes are available: tap-to-start, press-and-hold and a mixture of the before mentioned, called single-tap. The tap-to-start option lets the user tap it once to start and tap again to stop the recording. The press-and-hold option let the user record while pressing the microphone button. The last option is well suited for motor impaired patients, giving them a larger time frame to hold down the button and letting go, while still being able to record. The transcription can be read aloud via the device’s text-to-speech system with numerous voices. This enables the speaker to communicate in situation where the other party cannot read the screen.
The design makes it possible for the speaker to speak freely with another party, improving the quality of life significantly over text-based communication alternatives.
Another central feature of the app is its correction and feedback mechanism. When getting the transcription, the speaker can edit the recognized text directly if it contains errors. That text can then be sent back to the server and is saved together with the audio sample (see Figure \ref{fig:app}).

The speaker has also reported an improvement of its own voice because of the application. Being able to use the app as a fallback option when speech alone is insufficient for understanding. This has given the person confidence to speak more often.  

\section{Conclusion}

In this work, we presented a personalized ASR system that adapts a foundation model to the speech of a single dysarthric speaker. Starting from a baseline WER of 128.4\% / 109.7\%, which rendered the off-the-shelf model effectively unusable, full fine-tuning of Whisper reduced the WER to 15.8\% with only 1.4 hours of adaptation data and to 9.7\% when all collected data, including user corrections, is used and 7.8\% on the corrections test set. Our comparison showed that performance improves steadily with more adaptation data, with the largest gains occurring early and a favorable trade-off already reached at around 22.5 hours (10.7\% WER / 16.1\%). Full fine-tuning consistently outperformed LoRA adaptation, and Whisper proved a stronger foundation model than Qwen3-ASR for this speaker, suggesting that the substantial mismatch between dysarthric and standard speech benefits from adapting the full model capacity.

Beyond the recognition results, we deployed the adapted model in a mobile application designed around the speaker's accessibility needs, which both serves as a practical communication aid and enables continuous collection of corrected data through everyday use. The speaker reported a meaningful improvement in quality of life and increased confidence in speaking. Together, these results demonstrate that personalized fine-tuning can turn a foundation ASR model that fails on dysarthric speech into a system accurate and robust enough for real-world deployment.

\subsection{Limitations} 

Currently, the ASR model runs on a remote server, requiring an internet connection for use. Local on-device deployment would improve privacy and reliability, but real-time inference of large models
remains computationally challenging. 
For real-time on-device inference, quantization and / or the use of a smaller model might be necessary. However, this might degrade performance.

The current system is personalized to a single dysarthric speaker, making it unclear how much adaptation data would be required for other speakers with different profiles. Although our results demonstrate that fine-tuning is highly effective for our user, the approach
probably does not generalize to other speakers
without collecting
some
speaker-specific data, which represents a significant barrier for broader deployment. 

Finally, scaling this approach to many speakers simultaneously, potentially through a single multi-speaker adaptive model, remains an open research question we leave for future work.
Another goal we have set with our user has been using an Apple Watch companion app. This has been identified as a promising extension by the user, allowing a more convenient recording. The text and the resulting generated voice would still be played from the phone which can be left next to the user. This would allow the user to communicate without any devices in hand.

\section{Acknowledgment}

\ifreview
BLIND
\else
The projects on which this research is based were funded by
the Horizon research and innovation program of the European Union under grant agreement No 101135798 (Meetween) and 101213369 (DVPS),
and the KIT Campus Transfer GmbH (KCT) staff in accordance with the collaboration with Carnegie – AI.
The authors gratefully acknowledge the support.

The speaker approved the collection of the data and the use for research purposes and signed a corresponding form.
\fi

\section{Use of generative AI tools}
\ifreview
BLIND
\else
Generative AI tools were used in a limited capacity during the preparation of this work. Specifically, AI-assisted code completion was employed to support software development tasks. Language model suggestions were used to refine the clarity and style of the written text. Additionally, generative AI tools assisted in the enhancement of figures. All substantive intellectual contributions, including the research design, methodology, analysis, and conclusions, are entirely the authors' own.
\fi

\bibliographystyle{IEEEtran}
\bibliography{mybib}

\end{document}